\documentclass{article}
\usepackage{spconf,amsmath,epsfig}

\title{Portability of Syntactic Structure for Language Modeling}
\name{Ciprian Chelba}
\address{Microsoft Speech.Net / Microsoft Research\\
	One Microsoft Way\\
	Redmond, WA 98052\\
	chelba@microsoft.com
}

\begin{document}
%
\maketitle
\begin{abstract}
The paper presents a study on the portability of statistical syntactic knowledge
in the framework of the structured language model (SLM). We
investigate the impact of porting SLM statistics from the Wall Street
Journal (WSJ) to the Air Travel Information System (ATIS)
domain. We compare this approach to applying the Microsoft rule-based
parser (NLPwin) for the ATIS data and to using a small amount of data manually parsed at UPenn
for gathering the intial SLM statistics. 
Surprisingly, despite the fact that it performs modestly in perplexity (PPL), the model
initialized on WSJ parses outperforms the other initialization methods
based on in-domain annotated data, achieving a significant
0.4\% absolute and 7\% relative reduction in word error rate (WER) over a
baseline system whose word error rate is 5.8\%; the improvement measured
relative to the minimum WER achievable on the N-best lists we worked
with is 12\%.
\end{abstract}
\section{Introduction}
\label{sec:intro}

The structured language model uses hidden
parse trees to assign conditional word-level language model probabilities.
The model is trained in two stages: first the model parameters are intialized from a
treebank and then an N-best EM variant is employed for reestimating
the model parameters. 

Assuming that we wish to port the SLM to a new domain we have four alternatives for
initializing the SLM:\\
$\bullet$ manual annotation of sentences with parse structure. This is expensive,
time consuming and requires linguistic expertise. Consequently, only a
small amount of data could be annotated this way.\\
$\bullet$ parse the training sentences in the new domain using an automatic parser (\cite{charniak00}, \cite{mike:thesis},
\cite{ratnaparkhi:parser}) trained on a domain where a treebank is available already\\
$\bullet$ use a rule-based domain-independent parser (\cite{msft:nlpwin})\\
$\bullet$ port the SLM statistics as intialized on the treebanked-domain. Due to the way the
SLM parameter reestimation works, this is equivalent
to using the SLM as an automatic parser trained on the
treebanked-domain and then applied to the new-domain training data. 

We investigate the impact of different intialization methods and whether one can port statistical
syntactic knowledge from a domain to another. The second training
stage of the SLM is invariant during the experiments presented here.

We show that one can successfuly port syntactic knowledge from the Wall Street
Journal (WSJ) domain --- for which a manual treebank~\cite{UPenn}
was developed (approximatively 1M words of text) --- to the Air Travel Information System (ATIS)~\cite{atis} domain.
The choice for the ATIS domain was motivated by the fact that it is
different enough in style and structure from the WSJ domain and there is a small
amount of manually parsed ATIS data (approximatively 5k words) which allows us to train the SLM
on in-domain hand-parsed data as well and thus make a more interesting
comparison.

The remaining part of the paper is organized as follows:
Section~\ref{sec:slm_overview} briefly describes the SLM followed by
Section~\ref{sec:experiments} describing the experimental setup and
results. Section~\ref{sec:conclusions} discusses the results and
indicates future research directions.
  
\section{Structured Language Model Overview}
\label{sec:slm_overview}
An extensive presentation of the SLM can be found
in~\cite{chelba00}. The model assigns a probability $P(W,T)$ to every
sentence $W$ and its every possible binary parse $T$. The
terminals of $T$ are the words of $W$ with POStags, and the nodes of $T$ are
annotated with phrase headwords and non-terminal labels.
\begin{figure}[h]
  \begin{center}
    \epsfig{file=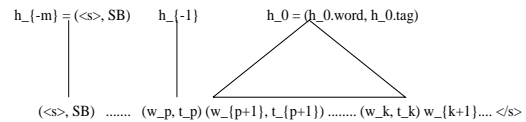,height=1.5cm,width=7cm}
  \end{center}
  \vspace{-0.5cm}
  \caption{A word-parse $k$-prefix} \label{fig:w_parse}
\end{figure}
 Let $W$ be a sentence of length $n$ words to which we have prepended
the sentence begining marker \verb+<s>+ and appended the sentence end
marker \verb+</s>+ so that $w_0 = $\verb+<s>+ and $w_{n+1} = $\verb+</s>+.
Let $W_k = w_0 \ldots w_k$ be the word $k$-prefix of the sentence ---
the words from the begining of the sentence up to the current position
 $k$ --- and  \mbox{$W_k T_k$} the \emph{word-parse $k$-prefix}. Figure~\ref{fig:w_parse} shows a
word-parse $k$-prefix; \verb|h_0 .. h_{-m}| are the \emph{exposed
 heads}, each head being a pair (headword, non-terminal label), or
(word,  POStag) in the case of a root-only tree. The exposed heads at
a given position $k$ in the input sentence are a function of the
word-parse $k$-prefix.

\subsection{Probabilistic Model} \label{ssec:prob_model}

\begin{figure}
  \begin{center} 
    \epsfig{file=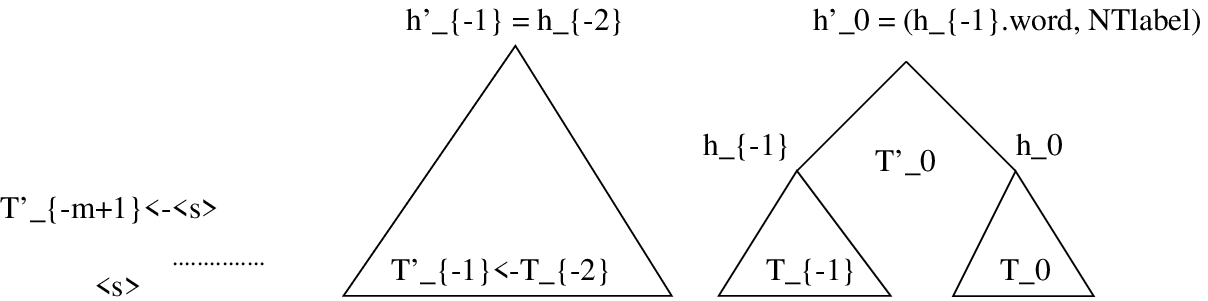,height=1.7cm,width=7cm}
  \end{center}
  \vspace{-0.5cm}
  \caption{Result of adjoin-left under NTlabel} \label{fig:after_a_l}
  \vspace{-0.5cm}
\end{figure}

\begin{figure}
  \begin{center} 
    \epsfig{file=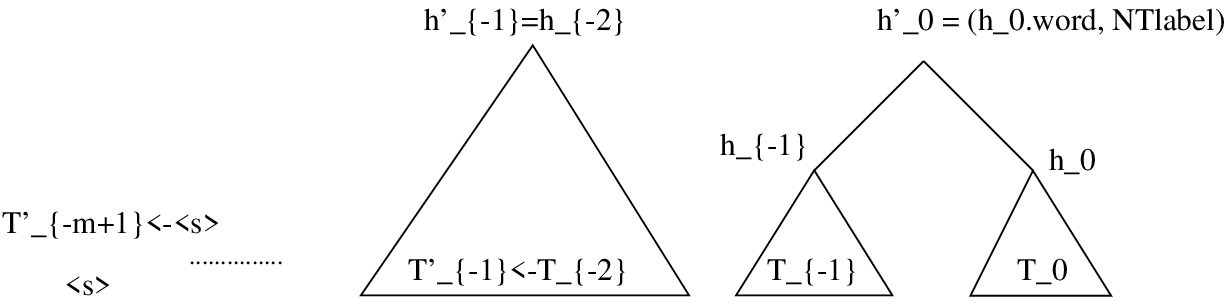,height=1.7cm,width=7cm}
  \end{center}
  \vspace{-0.5cm}
  \caption{Result of adjoin-right under NTlabel} \label{fig:after_a_r}
  \vspace{-0.5cm}
\end{figure}

 The joint probability $P(W,T)$ of a word sequence $W$ and a complete parse
$T$ can be broken into:
\begin{eqnarray}
\lefteqn{P(W,T)= } \nonumber\\
& \prod_{k=1}^{n+1}[&P(w_k/W_{k-1}T_{k-1}) \cdot P(t_k/W_{k-1}T_{k-1},w_k) \cdot \nonumber\\
&  & \prod_{i=1}^{N_k}P(p_i^k/W_{k-1}T_{k-1},w_k,t_k,p_1^k\ldots
p_{i-1}^k)] \label{eq:model}
\end{eqnarray}
where: \\
$\bullet$ $W_{k-1} T_{k-1}$ is the word-parse $(k-1)$-prefix\\
$\bullet$ $w_k$ is the word predicted by WORD-PREDICTOR\\
$\bullet$ $t_k$ is the tag assigned to $w_k$ by the TAGGER\\
$\bullet$ $N_k - 1$ is the number of operations the PARSER executes at 
sentence position $k$ before passing control to the  WORD-PREDICTOR
(the $N_k$-th operation at position k is the \verb+null+ transition);
$N_k$ is a function of $T$\\
$\bullet$ $p_i^k$ denotes the $i$-th PARSER operation carried out at
position k in the word string; the operations performed by the
PARSER are illustrated in
Figures~\ref{fig:after_a_l}-\ref{fig:after_a_r} and they ensure that
all possible binary branching parses with all possible headword and
non-terminal label assignments for the $w_1 \ldots w_k$ word
sequence can be generated. The $p_1^k \ldots p_{N_k}^k$ sequence of PARSER
operations at position $k$ grows the word-parse $(k-1)$-prefix into a
word-parse $k$-prefix.
 
Our model is based on three probabilities, each estimated using deleted
interpolation and parameterized (approximated) as follows:
\begin{eqnarray}
  P(w_k/W_{k-1} T_{k-1}) & = & P(w_k/h_0, h_{-1})\label{eq:1}\\
  P(t_k/w_k,W_{k-1} T_{k-1}) & = & P(t_k/w_k, h_0, h_{-1})\label{eq:2}\\
  P(p_i^k/W_{k}T_{k}) & = & P(p_i^k/h_0, h_{-1})\label{eq:3}
\end{eqnarray}%
 It is worth noting that if the binary branching structure
developed by the parser were always right-branching and we mapped the
POStag and non-terminal label vocabularies to a single type then our
model would be equivalent to a trigram language model.
 Since the number of parses  for a given word prefix $W_{k}$ grows
exponentially with $k$, $|\{T_{k}\}| \sim O(2^k)$, the state space of
our model is huge even for relatively short sentences, so we had to use
a search strategy that prunes it. Our choice was a synchronous
multi-stack search algorithm which is very similar to a beam search.

The \emph{language model} probability assignment for the word at position $k+1$ in the input
sentence is made using:
\begin{eqnarray}
P_{SLM}(w_{k+1}/W_{k}) & =
& \sum_{T_{k}\in
  S_{k}}P(w_{k+1}/W_{k}T_{k})\cdot\rho(W_{k},T_{k}),\nonumber \\ 
\rho(W_{k},T_{k}) & = & P(W_{k}T_{k})/\sum_{T_{k} \in S_{k}}P(W_{k}T_{k})\label{eq:ppl1}
\end{eqnarray}
which ensures a proper probability over strings $W^*$, where $S_{k}$ is
the set of all parses present in our stacks at the current stage $k$.

\subsection{Model Parameter Estimation} \label{ssec:model_estimation}

Each model component --- WORD-PREDICTOR, TAGGER, PARSER ---
is initialized from a set of parsed sentences after undergoing
headword percolation and binarization. 
Separately for each model component we:\\
$\bullet$ gather counts from ``main'' data --- about 90\% of the
training data\\
$\bullet$ estimate the interpolation coefficients on counts gathered from
``check'' data --- the remaining 10\% of the training data.

An N-best EM~\cite{em77} variant is then employed to jointly
reestimate the model parameters such that the PPL on training data is
decreased --- the likelihood of the training data under our model is
increased. The reduction in PPL is shown experimentally to carry over
to the test data.


\section{Experiments}
\label{sec:experiments}

We have experimented with three different ways of gathering the
initial counts for the SLM --- see Section~\ref{ssec:model_estimation}:\\
$\bullet$ parse the training data (approximatively 76k words) using Microsoft's
NLPwin and then intialize the SLM from these parse
trees. NLPwin is a rule-based domain-independent parser developed by
the natural language processing group at Microsoft~\cite{msft:nlpwin}.\\
$\bullet$ use the limited amount of manually parsed ATIS-3 data (approximatively 5k words)\\
$\bullet$ use the manually parsed data in the WSJ section
of the Upenn Treebank. We have used the 00-22
sections (about 1M words) for initializing the WSJ SLM. The word vocabulary used
for initializing the SLM on the WSJ data was the ATIS open vocabulary
--- thus a lot of word types were mapped to the unknown word type.

After gathering the initial counts for all the SLM model components as
described above, the SLM training proceeds in exactly the same way in
all three scenarios.
We reestimate the model parameters by training the SLM on
the \emph{same} training data (word level information only, all parse
annotation information used for intialization is ignored during this stage), namely
the ATIS-3 training data (approximatively 76k words), and using the
\emph{same} word vocabulary. Finally, we
interpolate the SLM with a 3-gram model estimated using deleted interpolation:
$$P(\cdot)=\lambda \cdot P_{3gram}(\cdot) +(1-\lambda) \cdot P_{SLM}(\cdot)$$
For the word error rate (WER) experiments we used the 3-gram scores assigned by the
baseline back-off 3-gram model used in the decoder whereas for the
perplexity experiments we have used a deleted interpolation 3-gram
built on the ATIS-3 training data tokenized such that it matches the
UPenn Treebank style.

\subsection{Experimental Setup}
\label{ssec:exp_setup}

The vocabulary used by the recognizer was re-tokenized such that it
matches the Upenn vocabulary --- e.g.~\emph{don't} is changed to \emph{do n't}, see~\cite{chelba00} for an accurate
description. The re-tokenized vocabulary size was 1k.  The size of the
test set was 9.6k words. The OOV rate in the test set relative to the
recognizer's vocabulary was 0.5\%.

The settings for the SLM parameters were kept constant accross all
experiments to typical values --- see~\cite{chelba00}. The
interpolation weight between the SLM and the 3-gram model was
determined on the check set such that it minimized the perplexity of
the model initialized on ATIS manual parses and then fixed for the
rest of the experiments.

For the speech recognition experiments we have used N-best hypotheses
generated using the Microsoft Whisper speech
recognizer~\cite{msft:whisper} in a standard setup:\\
$\bullet$ feature extraction: MFCC with energy, one and two adjiacent frame
differences respectively. The sampling frequency is 16kHz.\\
$\bullet$ acoustic model: standard senone-based, 2000 senones, 12 Gaussians per
mixture, gender-independent models\\
$\bullet$ language model: Katz back-off 3-gram trained on the ATIS-3 training
data (approximatively 76k words)\\
$\bullet$ time-synchronous Viterbi beam search decoder

The N-best lists (N=30) are derived by performing an $A^*$ search
on the word hypotheses produced by the decoder during the search for
the single best hypothesis. The 1-best WER ---baseline --- is 5.8\% .
The best achievable WER on the N-best lists generated this way is
2.1\% --- ORACLE WER --- and is the lower bound on the
SLM performance in our experimental setup. 

\subsection{Perplexity results}
\label{ssec:ppl}

The perplexity results obtained in our experiments are summarized
in Table~\ref{tab:ppl_results}.
Judging on the initial perplexity of the stand-alone SLM
($\lambda=0.0$), the best way to intialize the SLM seems to be by
using the NLPwin parsed data; the meager 5k words of manually parsed
data available for ATIS leads to sparse statistics in the SLM and the
WSJ statistics are completely mismatched. However, the SLM iterative training
procedure is able to overcome both these handicaps and after 13 iterations we
end up with almost the same perplexity --- within 5\% relative of the
NLPwin trained SLM but still above the 3-gram
performance. Interpolation with the 3-gram model brings the perplexity
of the trained models at roughly the same value, showing an overall modest 6\% reduction in perplexity over
the 3-gram model.
\begin{table}[htbp]
  \begin{center}
    \begin{tabular}{|l|l|r|r|r|}\hline
      Initial Stats & Iter			& $\lambda$ = 0.0 & $\lambda$ = 0.6 & $\lambda$ = 1.0 	\\ \hline
      NLPwin parses & 0 			& 21.3	& 16.7	& 16.9 	\\
      NLPwin parses & 13			& 17.2	& 15.9	& 16.9	\\ \hline
      SLM-atis parses & 0 			& 64.4	& 18.2	& 16.9 	\\
      SLM-atis parses & 13			& 17.8	& 15.9	& 16.9	\\ \hline
      SLM-wsj parses & 0 			& 8311	& 22.5	& 16.9 	\\
      SLM-wsj parses & 13			& 17.7	& 15.8	& 16.9	\\ \hline
    \end{tabular}
    \caption{Deleted Interpolation 3-gram $+$ SLM; PPL Results}
    \label{tab:ppl_results}
  \end{center}
  \vspace{-0.5cm}
\end{table}

One important observation that needs to be made at this point is that
although the initial SLM statistics come from different amounts of
training data, all the models end up being trained on the same number
of words --- the ATIS-3 training data. Table~\ref{tab:no_params}
shows the number of distinct types (number of parameters) in the
PREDICTOR and PARSER (see Eq.~\ref{eq:1}~and~\ref{eq:3}) components of the SLM in each training
scenario. It can be noticed that the models end up having roughly the
same number of parameters (iteration 13) despite the vast
differences at initialization (iteration 0).
\begin{table}[htbp]
  \begin{center}
    \begin{tabular}{|l|l|r|r|}\hline
      Initial Stats & Iter			& PREDICTOR		& PARSER	\\ \hline
      NLPwin parses & 0 			& 23,621		& 37,702	\\
      NLPwin parses & 13			& 58,405		& 83,321	\\ \hline
      SLM-atis parses & 0 			& 2,048			& 2,990		\\
      SLM-atis parses & 13			& 52,588 		& 60,983	\\ \hline
      SLM-wsj parses & 0 			& 171,471		& 150,751	\\
      SLM-wsj parses & 13			& 58,073		& 76,975	\\ \hline
    \end{tabular}
    \caption{Number of parameters for SLM components}
    \label{tab:no_params}
  \end{center}
  \vspace{-0.5cm}
\end{table}

\subsection{N-best rescoring results}
\label{ssec:wer}

We have evaluated the models intialized in different conditions
in a two pass --- N-best rescoring --- speech recognition setup. As
can be seen from the results presented in Table~\ref{tab:wer_results}
the SLM interpolated with the 3-gram performs best. The SLM
reestimation does not help except for the model initialized on the
highly mismatched WSJ parses, in which case it proves extremely
effective in smoothing out the SLM component statistics coming from
out-of-domain. Not only is the improvement from the mismatched initial
model large, but the trained SLM also outperforms the baseline and the
SLM initialized on in-domain annotated data. We attribute this
improvement to the fact that the initial model statistics on WSJ were estimated on a
lot more data (more reliable) than the statistics coming from
the little amount of ATIS data.

The SLM trained on WSJ parses achieved 0.4\% absolute
and 7\% relative reduction in WER over the 3-gram baseline of 5.8\%. 
The improvement relative to the minimum --- ORACLE
--- WER achievable on the N-best list we worked with is in fact 12\%.
\begin{table}[htbp]
  \begin{center}
    \begin{tabular}{|l|l|r|r|r|}\hline
      Initial Stats & Iter			& $\lambda$ = 0.0 & $\lambda$ = 0.6 & $\lambda$ = 1.0 	\\ \hline
      NLPwin parses & 0 			& 6.4	& 5.6	& 5.8 	\\
      NLPwin parses & 13			& 6.4	& 5.7	& 5.8	\\ \hline
      SLM-atis parses & 0 			& 6.5	& 5.6	& 5.8 	\\
      SLM-atis parses & 13			& 6.6	& 5.7	& 5.8	\\ \hline
      SLM-wsj parses & 0 			& 12.5	& 6.3	& 5.8	\\
      SLM-wsj parses & 13			&  6.1	& \underline{5.4}	& 5.8	\\ \hline
    \end{tabular}
    \caption{Back-off 3-gram $+$ SLM; WER Results}
    \label{tab:wer_results}
  \end{center}
  \vspace{-0.5cm}
\end{table}
We have evaluated the statistical significance of the best result
relative to the baseline
using the standard test suite in the SCLITE package provided by
NIST. The results are presented in Table~\ref{tab:sig_results}.
We believe that for WER statistics the most relevant significance test
is the Matched Pair Sentence Segment one under which the SLM interpolated
with the 3-gram is significant at the 0.003 level.
\begin{table}[htbp]
  \begin{center}
    \begin{tabular}{|l|r|}\hline
	Test Name & p-value \\\hline
	Matched Pair Sentence Segment (Word Error) & 0.003 \\
	Signed Paired Comparison (Speaker WER) & 0.055 \\
	Wilcoxon Signed Rank (Speaker WER) & 0.008 \\
	McNemar (Sentence Error) & 0.041\\ \hline
    \end{tabular}
    \caption{Significance Testing Results}
    \label{tab:sig_results}
  \end{center}
  \vspace{-0.5cm}
\end{table}

\section{Conclusions}
\label{sec:conclusions}

The main conclusion that can be drawn is that the method for
initializing the SLM is very important to the performance of the
model. We consider this to be a promising venue for future research. The parameter reestimation
technique proves extremely effective in smoothing the statistics
coming from a different domain --- mismatched initial
statistics. 

The syntactic knowledge embodied in the SLM statistics is portable but
only in conjunction with the SLM parameter reestimation technique. The
significance of this result lies in the fact that it is possible
to use the SLM on a new domain where a treebank (be it generated
manually or automatically) is not available.

\section{Acknowledgements}
Special thanks to Xuedong Huang and Milind Mahajan for useful
discussions that contributed substantially to the work presented in
this paper and for making available the ATIS N-best
lists. Thanks to Eric Ringger for making available the NLPwin parser and
making the necessary adjustments on its functionality. 

\bibliographystyle{IEEEbib}
\bibliography{camera_ready}

\end{document}